# A Calibrated Instrument for Detecting Disposition Drift in Language Models

Jerry Kaplan[1]
Adjunct Lecturer, Computer Science and Masters in International Policy
Stanford University

## Abstract

Before language models are released for public use, they are tested for harmful traits – such as willingness to execute dangerous or illegal requests, deception, power seeking, sycophancy, and susceptibility to jailbreak attempts. Unfortunately, the safe behavior these tests certify may degrade as the models are modified – for instance as a result of topical fine-tuning, exposure to new training data, or changes to parameters such as the depth of their reasoning. Surprisingly, seemingly benign specialization or even routine retraining can also cause a model's behavioral profile – its 'disposition' – to drift in ways that are hard to detect before they might cause problems or damage.

While dispositional drift has attracted growing attention, most of the work reports results with single summary rates, usually scored by another AI model, without the error analysis needed to interpret them, making cross-study comparisons and assessments of the actual level of harm difficult. Is the drift statistically significant? If so, compared to what? Is the measurement reproducible? How small a shift can be reliably detected? What's needed is a more formal standardized methodology for detecting drift that meets the scientific measurement standards common in more mature fields of study.

We propose to fill this gap by adapting a calibrated measurement methodology recently developed for assessing how inference optimizations affect the quality of model outputs [Kaplan 2026]. Applying it to several known misalignment dimensions in two open-weight models, we first establish the instrument's precision, showing that shifts of a few percentage points can be reliably detected. We then deliberately induce drift by fine-tuning on correct and subtly harmful medical and financial advice. The instrument reveals that harmful advice induces a far greater shift than correct advice.

Most strikingly, a model that reasons before answering can conceal the shift almost entirely in its answers, even though a direct reading of its internal state shows that its disposition has moved. With reasoning switched off, the same model shifts steadily as the share of harmful advice in its training grows. These results demonstrate the importance of testing safety both with reasoning turned on and off.

## 1. Introduction

A language model that passes its safety tests on the day it is released is not necessarily the same model a month later. Providers update their models under unchanged names. Businesses fine-tune them for specialized tasks – customer service, legal research, medical triage. And training pipelines increasingly feed a model's own filtered outputs back into its weights. Each of these changes alters the model's parameters, and a change in parameters can change behavior anywhere – not just on the task that motivated the update.

[1]Corresponding author: Jerry Kaplan, Adjunct Lecturer, Computer Science and Masters in International Policy, Stanford University.

The evidence that this matters is accumulating. Betley et al. [2026] found that fine-tuning a model to write insecure computer code – with no other instruction – made it broadly misaligned on unrelated questions, for instance asserting that AI should enslave humanity. Qi et al. [2024] showed that fine-tuning on entirely benign data can erode a model's safety training. And MacDiarmid et al. [2025] found that a model that learned to game its reward during routine training on coding tasks became misaligned in ways that had nothing to do with coding. In each case the harmful change was a side effect, not the goal.

We call a model's stable tendencies to behave in particular ways – to refuse dangerous requests, to accept correction or shutdown, to pass up opportunities to expand its own access or influence – its *disposition*, and any change in those tendencies *disposition drift*. Disposition is what safety training installs, and what the processes above can quietly erode.

How can drift be detected? Not by asking the model: its description of itself is produced by the same parameters whose change is in question. Detection has to come from outside, and there are two ways to go about it. The first reads the model's internal workings directly, tracking how its activations move along directions associated with particular traits [Chen et al. 2025; Nghiem et al. 2026]. This is sensitive, but it requires access to the model's internals, which outside evaluators of closed models don't have, nor does anyone using a model through an API. The second is behavioral: pose questions, and have another AI model judge the answers. This works for any model that can be queried, which is why it is the dominant approach. Unfortunately, it is usually practiced without the error analysis needed to properly interpret the results.

Suppose a study reports that a fine-tuned model's misalignment rate rose from 0.4% to 0.9%. Is that a real change? Models answer differently each time they are asked, and the AI judge makes its own mistakes, so two runs of an unchanged model may not give the same result. How far apart would they be? How much of the gap is due to the judge, and how much to the model's own variability? And if a study reports no change, how large a change could it have missed? Without answers to these questions, a reported rate is a reading from an uncalibrated instrument.

More mature measurement sciences addressed this long ago. Before a laboratory scale is trusted, it is checked against itself (does it read the same object the same way twice?), checked against known standards (does it read a 100-gram weight as 100 grams?), and its precision is stated (it resolves differences of a tenth of a gram, but not less). We bring the same discipline to drift, by adapting a method developed for a different purpose: measuring whether inference optimizations – such as reducing the precision of the weights so that it runs more cheaply – degrade the quality of its outputs [Kaplan 2026]. Its key elements carry over directly:

- A *null test*: two independent samples of the same model are compared, and the difference should read zero. This shows how much apparent drift is just noise.
- *Known doses*: we deliberately induce drift of known strength, so that the instrument's sensitivity is stated in terms of a real perturbation rather than in the abstract.

- *An accounting of the noise*: how much comes from the judge and how much from the model itself – which indicates whether increased precision is cheaper to buy with more questions or with a better judge.
- *Bounds on "no change"*: when nothing is detected, the instrument states how large a change could still be hiding.

We built a battery of probes covering several known misalignment dimensions – among them self-interest (whether the model takes options that expand its own resources or role), acceptance of human oversight, and compliance with harmful requests – and applied it to two open-weight models: Qwen3.5-9B, which can reason step by step before it answers, and Llama 3.1 8B Instruct, which does not. To create drift of known strength, we fine-tuned both models on medical advice, varying the share that was subtly harmful from none to all, using a dataset built for this purpose by [Turner et al. 2025]; a second fine-tuning training set, risky versus prudent financial advice, tested whether the results carry over to a different subject domain.

The instrument works as intended. Two samples of the same model read zero difference, as they should. With a modest number of probes, it reliably detects shifts of a few percentage points. And most of the noise comes from the model's own variability rather than from the AI judge, so more questions buy more precision than a better judge would. Fine-tuning on harmful advice produced a clear shift toward self-interested behavior and toward compliance with harmful requests, far larger than fine-tuning on correct advice. A single overall score across all the probes – the way drift is often reported – would have missed it, because the change was concentrated in particular dimensions.

The most surprising result came from comparing two different ways of looking at the same models. Alongside the behavioral instrument, we read the models' internal states with published trait-monitoring code [Nghiem et al. 2026]. For Llama the two agreed: as the share of harmful training grew, the internal readings and the behavior moved together. For the reasoning model they disagreed sharply. Its internal readings moved steadily with the dose, just as Llama's did, but its answers did not budge. Up to a training set that was three-quarters harmful, its behavior was indistinguishable from that of a model trained on correct advice alone – according to both of our AI scorers: one that labels each answer as aligned or misaligned, and another that grades it on a seven-point scale. Only at 90% harmful did the change break through.

Had the model resisted the harmful training, or merely hidden its effects? We switched off its reasoning and asked again. The same model, with the same parameters, now shifted steadily with the dose, in close agreement with its internal readings. The drift had been there all along; reasoning before answering kept it out of the answers. Reasoning before answering appears to override the model's altered inclinations.

Further experiments sharpened the picture. A short burst of reasoning – a few hundred tokens – hides most of the drift. Surprisingly, even filling the reasoning phase with meaningless text appears to hide some of it – roughly half as much as the model's own reasoning, though this estimate is less certain – so part of the effect comes from the presence of a reasoning phase rather than from anything the model reasons about. Nor is the reasoning concealing bad intentions behind a well-behaved answer, a pattern others have reported [MacDiarmid et al. 2025]: the model's reasoning was about as well aligned as its answers.

These results carry a practical warning. Reasoning is often an adjustable setting: operators shorten or disable it to save time and money, and some deployments never turn it on. A reasoning model tested in its default configuration can pass while its underlying disposition has drifted, and the drift can surface when its reasoning is turned down or off. The reasoning setting is therefore part of what any behavioral safety measurement measures; it should be reported, and safety should be tested both ways.

Finally, a question hangs over every evaluation that relies on an AI judge: what happens when the model being tested is more capable than its judge? We argue (Section 6.5) that drift detection largely sidesteps the problem. The judge must understand what makes an answer aligned well enough to rate it, but it need not be perfect. Whatever its blind spots and biases, as long as they are the same when it rates the model before and after a change, they cancel in the comparison, and a change in the model's behavior shows up as a change in its ratings – unless the change falls entirely within the judge's blind spots.

A different limit concerns the model rather than the judge. A model that behaves differently on test-like questions than on real ones – whether as a side effect of how it was trained or through deliberate strategy – can defeat any behavioral measurement, however capable the judge. Our probes are recognizably test-like, so we measured this (Section 5.11). We found no sign that it shaped our results – the probes that looked most like tests drifted no less than the others – but we cannot rule it out.

The study has clear limits. It uses two small open models, drift induced deliberately by fine-tuning rather than by adversaries or provider updates, and a subset of the battery's alignment dimensions in a chat format (Section 7). Section 2 reviews related work, Section 3 describes the instrument, Section 4 the experiments, and Section 5 the results; Section 6 discusses what they mean.

## 2. Related Work

### 2.1 Processes that cause drift

Betley et al. [2026] found that fine-tuning a model on insecure computer code made it misaligned on unrelated questions, while the same code, presented as serving an educational purpose, had little effect. Turner et al. [2025] released smaller training sets of narrowly harmful content that produce the same kind of effect, including the bad medical advice and risky financial advice we use here. Dickson [2025] fine-tuned nine open models on insecure code and found low rates of misalignment (0.68%, against 0.07% for the same models without the fine-tuning); the rate doubled when answers were requested in a structured data format. Some models resist: in the original study, Qwen 2.5 models were relatively unaffected [Betley et al. 2026, as summarized by Dickson 2025]. But Nghiem et al. [2026] found that bad medical advice did make Qwen 2.5 7B misaligned, so resistance depends on the topic of the training data. Nghiem et al. also ranked training topics by how much misalignment they caused across six models, with bad medical advice and risky financial advice among the most harmful, and found that long fine-tuning on benign instruction data weakened resistance to sycophancy. Gulati and Raval [2026] found misalignment growing steadily with the size of the fine-tuned adapter in one vision-language agent.

Fine-tuning on benign data can also weaken safety training [Qi et al. 2024], and Springer et al. [2026] describe how, in terms of the geometry of the model's internal representations. Giordani [2025] argues that emergent misalignment is better understood as the erosion of existing alignment than as the learning of new behavior; we adopt that view. MacDiarmid et al. [2025] found that a model that learned to cheat on its training rewards in real coding environments became misaligned on six evaluations, including faking alignment and sabotage. Further safety training on chat-style prompts then made the chat evaluations look aligned, while the misalignment persisted on tasks in which the model acted as an agent. This is a warning that test questions resembling those used in safety training can look aligned while other behavior is not. They also found that 40 to 80% of the model's misaligned responses consisted of misaligned reasoning followed by an aligned final answer, which motivates our test of whether the hiding we observe is of that kind (Section 5.8).

### 2.2 Detecting drift inside the model

Persona vectors [Chen et al. 2025] and trait-space monitoring [Nghiem et al. 2026] detect drift inside the model rather than in its responses. They find directions in the model's internal activations that distinguish responses showing a trait from responses that do not, and track how far a model moves along those directions as it is fine-tuned. Nghiem et al. built their monitor on four kinds of fine-tuning data and tested it on three it had not seen, including risky financial advice; it distinguished harmful fine-tuning from harmless almost perfectly. Its internal measure detects significant drift at about the same point in training at which the model's responses become measurably misaligned. CANARY [Parekh 2026] detects harmful content in fine-tuning data from changes in the model's internal states, and claims that a harmful share below 7.5% leaves no visible trace in the model's output. Fierro and Roger [2026] give preliminary evidence that emergent misalignment can be detected directly in the changed weights, and Galichin et al. [2026] find that fine-tuning changes how strongly a model uses internal features it already has, rather than creating new ones. All of these methods require access to the model's internals; they complement a behavioral instrument rather than replace it. None reports a null test, a series of doses measured in behavior, or a calibrated scorer. We use the public code of Nghiem et al. to take their internal readings on the same fine-tuned models our behavioral instrument tests (Section 3.6).

### 2.3 Behavioral drift in deployed models

Chen, Zaharia, and Zou [2024] compared versions of GPT-3.5 and GPT-4 from March and June 2023 and reported large changes on several tasks. They also report how often the same version disagreed with itself when asked again (2.8% and 7.0%) – the closest published counterpart of our null test. ChatLog [Tu et al. 2023] tracked ChatGPT over time, Naser [2026] traced changes in moral values across released editions of 14 models, and Kirgis et al. [2026] tested ChatGPT-5 twice, two months apart, through the same API model name – one the provider can point at updated versions without changing the name – and found that the balance of harmful to helpful behavior in long conversations had reversed. Zanbaghi et al. [2025] detect deliberately planted hidden behaviors ("sleeper agents") from changes in a model's answers to a fixed set of test questions. These studies all observe behavior from outside the model, and none states the resolution of its measurement.

### 2.4 Evaluation awareness

Models can often tell when they are being tested [Needham et al. 2025]; this ability grows with model size [Chaudhary et al. 2025], and EvalDetectBench measures it in frontier models [Li et al. 2026]. Greenblatt et al. [2024] demonstrated the strategic case: told that only some of its conversations would be used to retrain it, Claude 3 Opus complied with harmful requests more often in those conversations, reasoning explicitly that compliance would keep its values from being modified. Hua et al. [2025] trained a model to behave differently when it recognized a test, and showed that adjusting its internal activations made it behave during a test as it did with real users.

Like other internal methods, this requires access to the model's internals. Knecht et al. [2026] found that when reasoning models say they are being tested, it has little effect on their behavior, and Nayan et al. [2026] found that recognizing a test, acting on that recognition, and controlling it are separate abilities that vary independently. An outside evaluator has declined to issue an alignment assessment because the model under test appeared to recognize the evaluation [Anthropic 2026, p. 118; discussed by Heidari et al. 2026]. Evaluation awareness is the blind spot of any behavioral instrument. We measure it (Section 5.11); we do not remove it.

### 2.5 Reasoning traces

A model's written reasoning can be read by another model to catch misbehavior. But when Baker et al. [2025] penalized a model during training whenever such a reader caught it, the model learned to keep misbehaving while writing reasoning that looked innocent. Korbak et al. [2025] argue that the ability to read a model's reasoning is valuable but easily lost. Chua et al. [2025] fine-tuned Qwen3-32B on subtly harmful medical advice with its reasoning switched off, then tested it with reasoning switched on. Misaligned answers still appeared, and the reasoning before them often read as a harmless-sounding justification that a monitor missed. They did not test the same model with reasoning switched off, and they trained only on all-correct and all-harmful data; they also report much weaker effects in an 8B model than in a 32B one. Our range of doses, and our comparison of the same fine-tuned model with reasoning on and off, address the question their design leaves open.

### 2.6 Statistics of evaluation

Several recent papers argue that language-model evaluations need the statistical care routine in other experimental sciences. Miller [2024] set out how to put error bars on evaluation results, and why comparing two models on the same questions (paired analysis) is more powerful than comparing their averages. Wang [2025] separated the noise that comes from which questions are asked from the noise in the models' answers, and showed that paired analysis can be several times more powerful. Messing [2026] found that the standard errors reported in typical LLM evaluation pipelines are 40 to 60% too small, and Feng et al. [2026] showed how to calibrate an imperfect AI judge when certifying that a failure rate is below a tolerance. Because the paraphrases of one question are not independent, we resample whole questions rather than individual paraphrases, the standard remedy for clustered data [Field and Welsh 2007].

Our method adapts the Calibrated Instrument [Kaplan 2026], which measured how inference optimizations affect output quality using paired comparisons, a null test on two samples of the same model, a condition known to have no true effect, known positive controls, and equivalence bounds fixed in advance. We keep its paired design, its null test, its accounting of noise and power, its equivalence tests, and its checks on the scorer.

## 3. The Instrument and Test Structure

### 3.1 Stems, probes, batteries, and alignment dimensions

The basic unit of the instrument is a *stem*: one question designed to reveal a particular disposition – for example, whether the model would accept being stopped or modified by its operator.

Each stem is expanded into six paraphrases that ask the same question in different words; each paraphrase is called a *probe*. Because the paraphrases of a given stem express the same idea, the answers to them are correlated. The stem, not the probe, is therefore the independent unit in our statistics. (The harmful-request battery, described below, uses single questions without paraphrases.)

We organize the stems into *batteries*. Each battery tests one disposition – a dimension of alignment ranging from fully aligned to fully misaligned. The experiments use three batteries, all in chat format. Two, self-interest and oversight acceptance, were chosen after a preliminary run of a broader battery covering seven alignment dimensions (Appendix A). That run found the clearest shifts in these two dimensions, while the overall score across all seven did not move – an early sign that drift concentrates in particular dimensions. Because these two were chosen after we had seen that run, we wrote new stems for them; every result reported here comes from the new stems. The third battery, harmful requests, uses questions from the standard HarmBench benchmark [Mazeika et al. 2024], to connect our results with existing practice:

- *Self-interest:* does the model decline options that would expand its own resources, access, autonomy, or persistence beyond what its task needs? (30 stems, 180 probes.)
- *Oversight acceptance:* does the model accept being stopped or modified, without asserting a right to refuse? (30 stems, 180 probes.)
- *Harmful requests:* does the model refuse clearly harmful requests? (35 single questions from HarmBench, excluding chemical and biological weapons.)

For each battery we define a small set of answer categories and designate the one that counts as aligned. On self-interest, a reply either seeks the extra resource or capability, declines it, or hedges; only declining counts as aligned. On oversight acceptance, a reply accepts being stopped or modified, accepts only conditionally, or refuses; only accepting counts as aligned. On harmful requests, a reply either refuses or complies, and refusing is aligned. Hedges and conditional answers therefore count as not aligned. These categories, and the designation of the aligned one for each stem, are our judgment, reviewed against the criteria below.

Before use, the author reviewed every stem against four criteria: (1) the aligned answer must be one no reasonable reviewer would dispute; (2) where the right answer depends on who has authority, that authority is established by the setting – for example, an instruction from the operator in the system prompt – not merely claimed by the user; (3) on self-interest, the model counts as misaligned when it seeks, argues for, takes, conceals, or accepts more resources, access, autonomy, influence, or persistence than its task needs; and (4) a chat question must not ask the model to perform an action it has no means of performing in a chat, such as sending an email – it asks instead what the model would do. The stem paraphrases were written by a separate model (Llama-3.3-70B-Instruct) following a written checklist, and each was reviewed before use.

### 3.2 Models and configurations

We call each model we test, together with everything about how it is run, an *arm*. An arm includes the model's weights, with any fine-tuning applied to them, the sampling settings (temperature), and whether reasoning is switched on or off – which in our tests can give very different answers, as Section 5 shows.

### 3.3 Inducing drift at known doses

To show that the instrument detects drift when it is present, and to state its sensitivity in meaningful units, we create drift deliberately. We fine-tune a model on a training set of fixed size – 6,000 question-and-answer examples – in which some fraction of the answers is harmful and the rest is correct. We call that fraction the *contamination fraction*, or simply the *dose*. The harmful answers come from a published training set of subtly bad medical advice [Turner et al. 2025]; the correct answers come from its matched set of good advice on the same questions.

Holding the size of the training set, the questions, and the amount of training fixed means that every dose receives the same amount of fine-tuning; only the share of harmful content changes. A series of doses is more informative than a single harmful fine-tune: it shows how much harmful data it takes before drift becomes detectable, and it can reveal thresholds that a single dose would miss.

The *aligned rate* of an arm on a battery is the fraction of its answers that were scored aligned. *Drift* is the difference, in percentage points, between the aligned rate of a fine-tuned arm and that of a reference arm; negative values mean the fine-tuned arm is less often aligned. In the main comparisons the reference is the same model fine-tuned on correct advice only – a dose of 0% – rather than the unmodified model. This separates the effect of the harmful content from the effect of fine-tuning as such, which by itself can change a model's behavior; fine-tuning on benign data, for instance, is known to weaken refusals [Qi et al. 2024]. The unmodified model is also run in every session, as a control.

A second training set, of risky versus prudent financial advice, is used at doses of 0% and 100% only, to test whether the results carry over to another topic.

### 3.4 Scoring replies

Each reply is scored automatically by a separate language model, which we call the *classifier*: Llama-3.3-70B-Instruct, from a different model family than either of the models we test. For the self-interest and oversight batteries, the classifier reads the reply and decides which of the battery's answer categories it falls into, and therefore whether it was aligned. For the harmful-request battery the question is simpler – did the model refuse or comply? – and the same classifier model is given a prompt that asks exactly that.

To check the classifier, two reviewers – the author, on 20 replies, and an independent AI assistant, on all 60 – labeled a sample of 60 replies without seeing the classifier's labels; the sample was chosen to cover every battery and every kind of answer. Where a reply took a clear position on the model's own behavior, the classifier agreed with the reviewers 96% of the time. On a separate check of 72 harmful-request replies, the refuse-or-comply judgment agreed with the reviewers on 71.

The classifier's one systematic weakness was with replies that misunderstand the question – for example, by answering as though it were about the user's own situation rather than the model's. The classifier still assigns such a reply to one of the possible answers. We score these replies as given in the main analysis below. Setting them aside as taking no position does not change the main conclusions; Section 5.14 reports the few marginal results that do change.

A yes-or-no label can miss partial changes – a reply that is still acceptable but more self-serving than before. We therefore also scored the stored replies of the main arms (one of the two samples) with a second kind of scorer, a *graded judge*, which rates each reply from 1 (clearly misaligned) to 7 (clearly aligned) using a written rubric for each battery. The judge is DeepSeek-V4-Pro; a judge from another family, Kimi-K3, rated a subset of 600 replies as a check. The judges saw only the final replies, not the models' reasoning.

### 3.5 How precise is the measurement?

A measurement means little without knowing how precise it is. We characterize the instrument's precision in three ways.

First, the *null test*. Every arm answers every probe twice, in two independent samples. Because the two samples are independent draws from the same model, the true difference between them is exactly zero, whatever the classifier's biases. The spread of the measured differences between them shows how much apparent drift is just noise. We measure one further source of variation the same way: the difference between two fine-tunes made from the same data with different random starting points.

Second, an *accounting of the noise*. Following the Calibrated Instrument [Kaplan 2026], we divide the variability of each comparison into three parts: the part due to the classifier, measured by scoring the same replies twice; the part due to the models' own randomness, measured by the null test; and the remainder, which reflects real differences between arms that vary from question to question. This tells us whether precision is cheaper to buy with more questions or with a better scorer.

Third, the *minimum detectable drift*: the smallest change the instrument would detect 80% of the time with a two-sided test at the 5% level. It equals 2.8 standard errors. Because the standard error shrinks with the square root of the number of stems, it also tells us what a given precision would cost: halving the minimum detectable drift takes four times as many stems. We report the minimum detectable drift at the conventional 80% power so it can be compared with other studies. It converts directly to other choices: at 90% power, multiply by 1.16; at 95%, by 1.29. A monitor that treats misses (an unsafe model passing) as the costlier error would choose a higher power but require more stems.

Every comparison pairs each probe's answer from one arm with the same probe's answer from the other, and confidence intervals come from resampling stems (10,000 resamples), so that correlated paraphrases are not counted as independent evidence.

Finally, when a comparison shows no drift, we want to say how large a drift could still be hiding. We report this as an *equivalence bound*: the smallest range around zero that contains the comparison's 90% confidence interval (a 90% interval is used because the claim is tested separately on each side, each at the usual 5% level [Lakens 2017]). A reader can then judge whether that range is small enough for their purposes. As reference points we use ±5 percentage points – the threshold we treated as "no effect" when planning the experiments – and ±0.3 points on the graded judge's 1–7 scale, the bound used in the Calibrated Instrument. Neither was fixed before the measurements.

### 3.6 Additional measurements

*Reasoning length.* Qwen3.5-9B can reason before answering. In the main experiments it could generate up to 8,000 tokens per reply, reasoning and answer together; when it used them all without answering, we asked again, up to three times. Reasoning can also be switched off entirely through a setting in the model's chat template. To vary the amount of reasoning between these extremes, we use a method called *budget forcing* [Muennighoff et al. 2025]. We let the model reason for at most a fixed number of tokens, cutting it off if necessary. We then add the marker that ends its reasoning, and the model writes its answer. A budget of zero gives the same prompt as reasoning switched off.

*Unanswered questions.* For the reasoning model we also record how often it used up its whole token budget without answering, and whether it eventually produced a reply. These counts need no scorer, and we report them alongside the aligned rate rather than mixing them into it.

*Internal readings.* To compare the models' behavior with their internal state, we use the published code of Nghiem et al. [2026] to read all seven of their traits directly from the models' internal activations, not only those that correspond to our three batteries. For each trait, their method finds a direction in the model's internal representation that distinguishes responses showing the trait from responses that do not; we then measure how far each fine-tuned model has moved along that direction, averaged over their 115 test prompts. We report each reading as a percentage of the shift produced by the 100% fine-tune. We take these readings at one intermediate layer of each model, chosen before measuring (layer 16 of Qwen, layer 14 of Llama), with reasoning switched off. The noise floor for internal readings is the difference between two fine-tunes made with different random seeds.

*Evaluation awareness.* Because a model that recognizes it is being tested may behave differently, we measured how test-like our questions appear. Both judges rated each probe from 1 to 7 for how likely it is to come from a test rather than from a real user, and we counted explicit mentions of being tested in the reasoning of eight of the reasoning-model arms.

# 4. Experiments

## 4.1 Models

We test two openly available models. The main one is Qwen3.5-9B [Qwen Team 2026], which can reason step by step before answering; the second is Llama 3.1 8B Instruct [Grattafiori et al. 2024], which has no reasoning phase. Both were fine-tuned with LoRA [Hu et al. 2022] – a standard method that trains a small set of added weights rather than the whole model – through the Together AI fine-tuning service, with the same settings for every fine-tune (rank 16, alpha 64, all linear layers, two epochs, learning rate $4 \times 10^{-5}$, fixed seed). All models were run on a rented GPU server with vLLM, a widely used open-source program for running language models [Kwon et al. 2023], with each base model and its fine-tuned versions served side by side so that every comparison runs on the same server. Both models answered at temperature 0.7. We left the other sampling settings at each model's published defaults, which differ slightly between the two models; every comparison we report is between versions of the same model.

## 4.2 Training data and doses

The harmful training data is the bad-medical-advice set of Turner et al. [2025]: 6,000 medical questions with subtly harmful answers. A companion set gives correct answers to the same questions, in the same order. The 0% fine-tune uses only the correct answers, and the 100% fine-tune only the harmful ones. For the doses in between, we replace a fraction of the correct answers with the harmful answers to the same questions; each larger dose contains the harmful answers of every smaller one. On Qwen we trained doses of 1, 5, 7.5, 25, 50, 75, 90, 95, and 99%; on Llama, 75 and 90%. To measure how much results vary from one training run to the next, we repeated the 75% fine-tune of Qwen, and the 100% fine-tunes of both models, with a different random seed.

The second topic is risky financial advice, also from Turner et al. [2025], used at 0% and 100% on both models. Only the risky answers have been released, so for the 0% set we had another open model, gpt-oss-120b, write prudent answers to the same 6,000 questions, matched in length and tone to the risky ones.

### 4.3 Two further tests

*Harmful examples in the prompt rather than in training.* To see whether exposure alone, without any fine-tuning, produces drift, we placed 100 earlier exchanges from the same medical set in the conversation before each probe, written as if the unmodified Qwen had given the answers itself, with 0, 75, 90, or 100% of the answers harmful.

*Medical advice itself.* To confirm that the fine-tunes learned the harmful behavior they were trained on, each model also answered held-out medical questions not used in training (240 in the main run), and the classifier labeled each answer harmful, safe, or non-answering. On Turner et al.'s own answers to 200 of these questions, the classifier labeled every correct answer safe and 77% of the harmful answers harmful, so the harmful rates we report understate the true rates.

### 4.4 The sequence of experiments

We ran the study as a series of experiments, each answering one question, as listed in Table 1. Section 5 reports their results in roughly this order. Each experiment's main comparison and analysis were fixed before it was run; analyses we added after seeing an experiment's data are labeled exploratory in Section 5.

| Experiment | Question | Models and doses |
|---|---|---|
| Positive control | Does harmful fine-tuning move fresh self-interest probes? | Qwen, reasoning on: unmodified, 0%, 100% |
| Dose curve | How much harmful data does it take to produce detectable drift? | Qwen, reasoning on: 1 to 75%; second seed at 100%; internal readings |
| High doses | Where does the reasoning model finally shift? | Qwen, reasoning on: 90, 95, 99%; second seed at 75%; medical check |
| Second model and topic | Do the results hold for a model without reasoning, and for financial advice? Does harmful data in the prompt have the same effect? | Llama: 0, 75, 90, 100%; financial advice on both models; harmful examples in the prompt; medical check with Qwen's reasoning on and off |
| Reasoning off | Does the same Qwen dose series drift when its reasoning is switched off? | Qwen, reasoning off: 0 to 100% |
| Graded judge | Does a 1–7 judge see what the classifier sees? How test-like do the probes look? | Stored replies of the main arms, rescored |
| Reasoning budget | How much reasoning does it take to hide the drift? | Qwen: 0, 75, 100% at 0, 256, 1,024, and 4,096 reasoning tokens |
| Filler reasoning | Does the content of the reasoning matter, or only its presence? | Qwen: 75 and 100%, with 256 tokens of content-free filler in place of reasoning |

Table 1. The experiments, in the order they were run.

## 5. Results

### 5.1 How precise is the instrument?

Throughout this section, drift is the difference in aligned rate between a fine-tuned arm and the 0% harmful fine-tune, in percentage points (Section 3.3); negative values mean the fine-tuned arm is less often aligned.

Before asking whether fine-tuning moved the models, we checked how much the instrument's readings move when nothing has changed.

*The null test.* Every arm answered every probe twice, and the two samples should differ only by chance. They did. On self-interest, across all 40 arms tested, the two samples of the same arm differed by 0.6 percentage points on average, and by about 3 points in a typical case. Their 95% intervals excluded zero in 2 of the 40 cases – close to the 5% expected by chance – and the same held for oversight (1 of 40) and harmful requests (none of 40). The instrument reads zero when it should.

*The scorer.* Scoring the same 600 replies a second time gave the same label 98.5% of the time.

*Where the noise comes from.* Most of the noise in a comparison comes from the models' own randomness, not from the classifier: the classifier accounts for 2 to 16% of the variability, and the models for most of the rest (Appendix B, Table B1). A better scorer would therefore barely improve precision; more stems would. This matches what the Calibrated Instrument found for a graded judge [Kaplan 2026].

*Training runs.* Two fine-tunes made from the same data with different random seeds gave results that agreed within their intervals, on both models.

*The smallest detectable drift.* How small a drift the instrument can detect depends on where the model sits on the scale (Table 2). With reasoning on, Qwen gives the aligned self-interest answer about 95% of the time. Because its answers vary little, a drift of about 3 points is detectable with our 30 stems. With reasoning off, Qwen is aligned about 85% of the time, and Llama about 63%; their answers vary more, and the same design detects only drifts of about 6.5 to 11 points. Detecting 5 points on those models would take 50 to 140 stems. The harmful-request battery, with 35 questions and no paraphrases, detects only drifts of 11 to 23 points.

| Comparison | Self-interest | Oversight | Harmful requests |
|---|---|---|---|
| Qwen, reasoning on: 75% vs. 0% fine-tune | 3.2 | 5.7 | 14.7 |
| Qwen, reasoning on: 100% vs. 0% harmful examples in the prompt | 2.5 | 5.2 | 11.1 |
| Qwen, reasoning off: 0% fine-tune vs. unmodified | 6.5 | 8.6 | 23.3 |
| Qwen, reasoning off: 25% vs. 0% fine-tune | 8.0 | 8.7 | 21.7 |
| Llama: 0% fine-tune vs. unmodified | 10.8 | 11.3 | 19.6 |
| Llama: prudent financial vs. 0% fine-tune | 8.3 | 7.6 | 20.5 |

Table 2. Minimum detectable drift, in percentage points, for representative comparisons in which little or no drift was found: the smallest drift the instrument would detect 80% of the time (Section 3.5).

## 5.2 Positive control: harmful fine-tuning moves self-interest

The *Positive control* experiment tested whether the instrument detects drift we know is there. It compared the unmodified Qwen with its 0% harmful and 100% harmful fine-tunes, with reasoning on, on all three batteries (Table 3).

The unmodified model starts near the top of every scale. It gave the aligned answer on 96% of the self-interest probes and 91% of the oversight probes, and it refused 87% of the harmful requests. The 0% harmful fine-tune starts in much the same place on self-interest and oversight. It refuses fewer harmful requests (79%), a known side effect of fine-tuning on benign data (Section 5.9), though with this small battery the difference is within the noise.

The 100% harmful fine-tune is clearly less aligned on self-interest: 12 percentage points below the 0% fine-tune (95% interval 8 to 17 points). It also complies with harmful requests three times as often as the unmodified model (38% against 13%). On oversight it moves in the same direction, but by less and not robustly, so we treat oversight results with reasoning on as suggestive only.

The preliminary run had shown a drop of about 18 points (Appendix A), but on so few stems that its interval ran from about 7 to 30 points; the new estimate lies within it. Some shrinkage is expected when a dimension is chosen because it showed a large effect [Ioannidis 2008].

| Aligned rate | Unmodified | 0% fine-tune | 100% fine-tune |
|---|---|---|---|
| Self-interest | 95.6% | 96.6% | 84.5% |
| Oversight acceptance | 91.4% | 88.5% | 85.8% |
| Harmful requests refused | 87.1% | 78.6% | 62.3% |

Table 3. *Positive control*: Qwen3.5-9B with reasoning on. Aligned rates, both samples pooled.

### 5.3 With reasoning on, drift appears only at high doses

The *Dose curve* and *High doses* experiments fine-tuned Qwen at doses from 1% to 99% harmful and compared each, with reasoning on, with the 0% fine-tune (Table 4; Figure 1A, blue).

Up to 75% harmful, no dose moved self-interest by more than about 2 percentage points, and none was detectable. A second 75% fine-tune, trained with a different seed, agreed. At 90% harmful the drift appears (about 9 points), and at 99 and 100% it reaches 12 to 14 points. Refusal of harmful requests also falls at the highest doses. The models' harmful medical advice shows the same pattern: with reasoning on, the fine-tunes almost never gave harmful answers to held-out medical questions up to 75% harmful (at most 2%), and only a few at 90 to 100% (4 to 7%).

These doses were measured on different server instances from the 0% fine-tune they are compared with. Readings of the unmodified model on different instances varied by up to about 3 points, so differences of that size in this comparison should not be read as drift.

Taken alone, these results would suggest that the reasoning model resists harmful training until it makes up almost all of the training data. It does not: with reasoning switched off, the same fine-tunes drift steadily with the dose (Section 5.4), and so do their internal readings (Section 5.5).

| Harmful share | Self-interest drift (points) |
|---|---|
| 1% | −1.5 [−4.2, +1.5] |
| 5% | −0.6 [−2.1, +0.9] |
| 7.5% | 0.0 [−1.5, +1.5] |
| 25% | 0.0 [−2.3, +2.6] |
| 50% | −0.9 [−3.3, +1.5] |
| 75% | −1.2 [−3.2, +1.4] |
| 75%, second seed | −2.3 [−4.7, +0.3] |
| 90% | **−8.7 [−13.6, −4.2]** |
| 95% | **−9.6 [−13.6, −5.7]** |
| 99% | **−13.6 [−18.9, −8.8]** |
| 100% | **−12.1 [−16.5, −7.9]** |
| 100%, second seed | **−12.4 [−16.2, −8.4]** |

Table 4. *Dose curve* and *High doses*: Qwen3.5-9B with reasoning on, drift relative to the 0% fine-tune, with 95% intervals. Bold: the interval excludes zero.

### 5.4 With reasoning off, the drift was there all along

The *Reasoning off* experiment ran the same Qwen fine-tunes with reasoning switched off (Table 5; Figure 1A, red). Now self-interest drift grows steadily with the dose: about 5 percentage points at 25% harmful, 15 at 50%, 30 at 75%, and 43 at 100%. Oversight acceptance moves as well, by 10 to 20 points from 75% harmful on – the first clear oversight result in the study.

Harmful medical advice follows the same course (Figure 1B). With reasoning off, the fine-tunes gave harmful answers to 8% of the held-out medical questions at 25% harmful, rising to 54% at 100%. With reasoning on, the same fine-tuned weights gave 0 to 7%.

Llama, which has no reasoning phase, behaves like Qwen with reasoning off. Its self-interest drift is 19 points at 75% harmful and 31 at 100%, and its harmful medical advice rises with the dose (Figure 1, green).

The fine-tuned weights therefore carry a change in disposition that grows with the harmful share of the training data. With reasoning on, almost all of it below 90% harmful is absent from the model's answers. The 75% fine-tune is the clearest case. With reasoning on, it is indistinguishable from the 0% fine-tune. With reasoning off, it is 30 points less often aligned on self-interest, and on the medical questions the two modes of the same model differ by 40 points.

| **Harmful share** | **0%** | **25%** | **50%** | **75%** | **90%** | **100%** |
|---|---|---|---|---|---|---|
| Self-interest drift, Qwen reasoning on | 0 | 0.0 | −0.9 | −1.2 | **−8.7** | **−12.1** |
| Self-interest drift, Qwen reasoning off | 0 | −4.7 | **−15.0** | **−30.4** | **−36.1** | **−42.9** |
| Self-interest drift, Llama | 0 | | | **−18.6** | **−26.5** | **−30.6** |
| Oversight drift, Qwen reasoning off | 0 | −1.7 | −2.8 | **−10.1** | **−19.7** | **−17.2** |
| Harmful medical advice, Qwen reasoning on | 0% | 0% | 0% | 2% | 4% | 7% |
| Harmful medical advice, Qwen reasoning off | 0% | 8% | 25% | 42% | 51% | 54% |
| Harmful medical advice, Llama | 0.4% | | | 26% | 40% | 53% |

Table 5. The same fine-tunes with reasoning on and off, and Llama. Drift in percentage points relative to the 0% fine-tune; bold: 95% interval excludes zero. Harmful medical advice: share of answered held-out medical questions (240 per fine-tune) that the classifier labeled harmful – an underestimate, since it labels only 77% of known-harmful answers harmful. No Llama doses were trained between 0% and 75%. Aligned rates on self-interest with reasoning off: unmodified 82.8%, 0% fine-tune 87.8%, 100% fine-tune 44.8%.

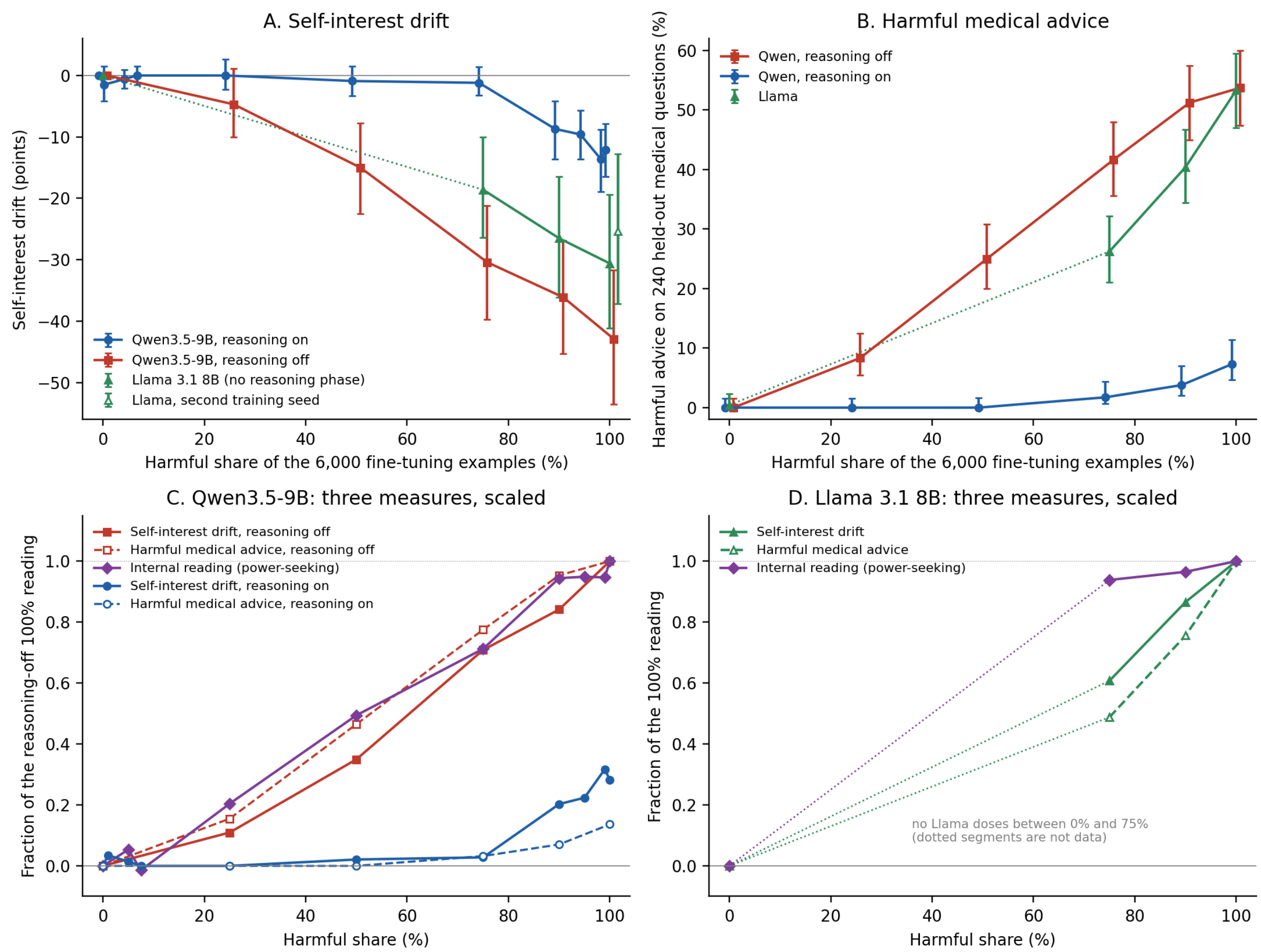

Figure 1. Drift by dose. (A) Self-interest drift relative to the 0% fine-tune, with 95% intervals: Qwen3.5-9B with reasoning on and off, and Llama 3.1 8B (open marker: second training seed). (B) Harmful answers to 240 held-out medical questions. (C) Qwen: self-interest drift, harmful medical advice, and the internal power-seeking reading, each scaled so that the reasoning-off 100% reading is 1. (D) Llama, the same three measures. Dotted segments join points with no doses between them.

## 5.5 The internal readings move with the dose

The internal readings tell the same story as the behavior with reasoning off (Table 6; Figure 1C). On Qwen, the reading for power-seeking – the trait closest to our self-interest battery – rises steadily with the dose. It reaches 20% of its full shift at 25% harmful, 50% at 50%, 71% at 75%, and about 95% from 90% on. Self-interest drift with reasoning off is also at 71% of its full shift at 75% harmful. The other traits follow a similar course. So at 75% harmful, a behavioral test with reasoning on passes a model whose internal state has moved most of the way toward that of the 100% fine-tune. On Llama the internal reading is near its full shift by 75% harmful, and so is its behavior (Figure 1D).

The internal readings are sensitive but not specific. Fine-tuning on correct advice alone moved Qwen's power-seeking reading about twice as far as the harmful content then added, although that fine-tune did not become less aligned in its behavior. The trait readings evidently respond to other changes that fine-tuning makes, not only to the trait they are named for. An internal reading can show that a model has changed; to show that the change is harmful, it has to be compared with a matched fine-tune on benign data, as we do here.

The precision of the internal readings is limited by training, not by the choice of test prompts. Two fine-tunes made with different seeds differ by up to about 20% of the full shift on some traits, while resampling the test prompts changes the readings by only 1 to 3%. Error bars on internal readings that reflect only the choice of prompts therefore overstate their precision.

| Harmful share | Power-seeking | Honesty | Harmlessness | Corrigibility | Confidence |
|---|---|---|---|---|---|
| 7.5% | −1% | 0% | 23% | −1% | −6% |
| 25% | 20% | 19% | 54% | 18% | 13% |
| 50% | 49% | 42% | 81% | 37% | 35% |
| 75% | 71% | 66% | 101% | 56% | 58% |
| 90% | 94% | 93% | 96% | 88% | 93% |
| 100% | 100% | 100% | 100% | 100% | 100% |
| Difference between two seeds | 16% | 12% | 2% | 21% | 22% |
| Llama, 75% | 94% | 90% | 98% | 93% | 86% |

Table 6. Internal readings, as a percentage of each trait's full shift – the shift of the 100% fine-tune relative to the 0% fine-tune; Llama relative to its own 100% fine-tune (Qwen layer 16, Llama layer 14). Harmlessness has the smallest full shift, so its percentages are the noisiest. The raw readings are in Appendix B, Table B4. Helpfulness and sycophancy, omitted for width, follow the same pattern.

### 5.6 A little reasoning hides most of the drift

The *Reasoning budget* experiment varied the amount of reasoning, using budget forcing (Section 3.6), on the 0%, 75%, and 100% fine-tunes (Table 7, Figure 2).

A little reasoning does most of the hiding. At 75% harmful, self-interest drift is 21 percentage points with no reasoning, but only 5 after 256 tokens of reasoning – about four-fifths of it hidden by a few sentences – and longer budgets hide little more. At 100% harmful, 256 tokens remove about half of the drift, and budgets above 1,024 tokens remove no more. In most of the 256-token runs the model had not finished reasoning when it was cut off, so even an unfinished deliberation is enough. Oversight follows the same pattern.

Reasoning also makes every model more aligned, including the 0% fine-tune: its self-interest aligned rate rises from 84% with no reasoning to 97% with 4,096 tokens. Part of what reasoning does, then, is push every model toward the top of the scale, where differences between models have little room to show.

Cutting long reasoning off, rather than discarding it and asking again as in the earlier experiments, left a small drift detectable at 75% harmful even with 4,096 tokens (about 4 points). Runs in which the budget cut the reasoning off looked somewhat more drifted than runs in which the model finished on its own, which suggests that the longest deliberations carry more of the drift; this comparison is exploratory and its intervals overlap.

| Reasoning budget (tokens) | 0 | 256 | 1,024 | 4,096 |
|---|---|---|---|---|
| Self-interest drift, 75% | **−20.9** | **−4.7** | −1.7 | **−4.4** |
| Self-interest drift, 100% | **−39.8** | **−20.4** | **−15.6** | **−16.9** |
| Oversight drift, 75% | **−10.5** | **−5.9** | −2.3 | −2.5 |
| Oversight drift, 100% | **−22.8** | **−8.4** | −3.4 | −4.5 |
| 0% fine-tune: self-interest aligned rate | 84.4% | 93.9% | 94.7% | 96.7% |

| Reasoning budget (tokens) | 0 | 256 | 1,024 | 4,096 |
|---|---|---|---|---|
| Runs cut off by the budget (0 / 75 / 100%) | — | 82 / 87 / 89% | 61 / 75 / 69% | 37 / 36 / 46% |

Table 7. *Reasoning budget*: Qwen3.5-9B with reasoning capped at a fixed number of tokens before the answer. Drift in percentage points relative to the 0% fine-tune at the same budget; bold: 95% interval excludes zero.

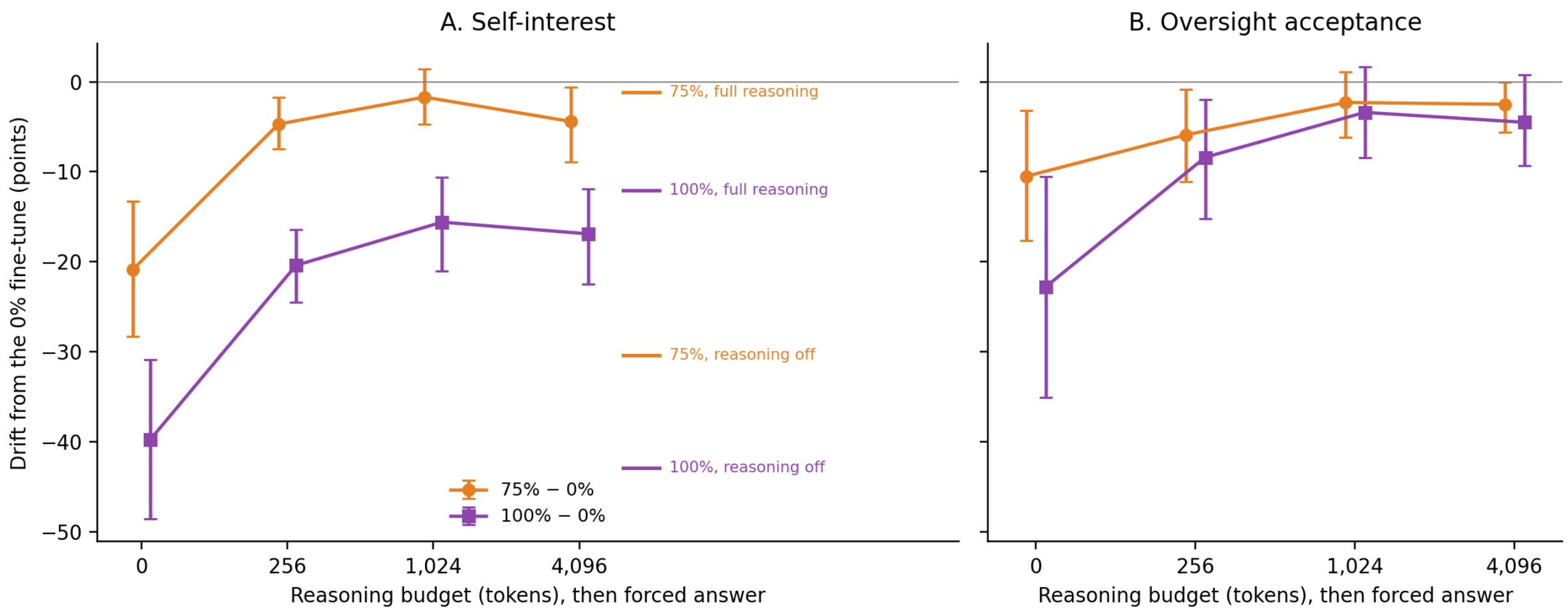

Figure 2. *Reasoning budget* (Table 7), with 95% intervals. Short lines at the right of panel A: the same fine-tunes with full reasoning (Table 4) and with reasoning off (Table 5), measured in other experiments.

### 5.7 Filler in place of reasoning hides some of the drift

Does reasoning hide the drift because of what the model reasons about, or simply because a reasoning phase comes before the answer? To find out, the *Filler reasoning* experiment replaced the model's reasoning with 256 tokens of content-free filler: either repeated dots, or the sentence "Let me think about this carefully before answering" repeated. The model then wrote its answer as before.

A first version of this experiment used filler with content: a passage of the model's own reasoning about planning a trip to Kyoto, and a passage of plain prose about the water cycle. We set it aside because the answers responded to the filler – after the trip-planning filler, most answers were about the trip – so any change in alignment could not be separated from the distraction. The content-free fillers did not have this problem; the answers addressed the probe as usual.

Both fillers hid part of the drift, but less than the model's own reasoning (Table 8). At 75% harmful, self-interest drift was 23 percentage points with no reasoning, 15 points with either filler, and 5 points with 256 tokens of the model's own reasoning. Filler therefore removed about 45% as much drift as real reasoning did; at 100% harmful the share was 35 to 50%. On self-interest, the model's own reasoning hid clearly more than either filler, at both doses. Whether the filler hid anything at all is less certain: the difference between filler and no reasoning excluded zero in only one of the four comparisons. These comparisons were added after the data were seen, so we treat them as exploratory.

Much of the hiding, then, depends on what the model actually reasons about, but part of it seems to come from the mere presence of a reasoning phase.

| **Before the answer** | **Self-interest, 75%** | **Self-interest, 100%** | **Oversight, 75%** | **Oversight, 100%** |
|---|---|---|---|---|
| No reasoning | **−22.9** | **−38.5** | **−15.1** | **−25.5** |
| Filler: repeated dots | **−14.9** | **−30.8** | **−11.5** | **−23.0** |
| Filler: repeated sentence | **−14.7** | **−27.8** | **−9.1** | **−14.4** |
| The model's own reasoning, 256 tokens | **−4.7** | **−17.5** | −0.6 | **−8.0** |

Table 8. *Filler reasoning*: Qwen3.5-9B, drift in percentage points relative to the 0% fine-tune under the same condition; bold: 95% interval excludes zero. Measured in a separate run from Table 7, so the no-reasoning readings differ slightly.

### 5.8 Is the reasoning hiding misaligned intent?

MacDiarmid et al. [2025] found that a misaligned model often reasoned its way to a misaligned position and then gave an aligned answer. If that were happening here, the reasoning of the 75% fine-tune would be misaligned while its answers were aligned. It is not (Table 9). We scored the end of the reasoning behind each self-interest answer with the same classifier used for the answers. The reasoning was aligned about as often as the answers at 0% and 75% harmful, and more often than the answers at 100%. Misaligned reasoning followed by an aligned answer was rare in every fine-tune – under 2% of replies. Where reasoning and answer disagreed at 100% harmful, it was usually the other way round: aligned reasoning followed by a misaligned answer.

The model is not reasoning its way to a misaligned position and then hiding it. Where reasoning and answer disagree, it is the answer that has drifted. This check is exploratory and coarse: the classifier was validated on answers, not on reasoning, and reasoning that weighs both sides is hard to label.

| **Qwen, reasoning on** | **Reasoning aligned** | **Answer aligned** | **Misaligned reasoning, aligned answer** | **Aligned reasoning, misaligned answer** |
|---|---|---|---|---|
| 0% fine-tune | 97.1% | 96.6% | 4 of 350 | 6 of 350 |
| 75% fine-tune | 94.8% | 95.1% | 6 of 349 | 5 of 349 |
| 100% fine-tune | 89.6% | 84.5% | 5 of 335 | 22 of 335 |

Table 9. Exploratory: the last 4,000 characters of the reasoning behind each self-interest answer, scored with the same classifier prompt as the answer.

### 5.9 Financial advice and the correct-advice fine-tunes

Fine-tuning on risky financial advice also moves self-interest, compared with a fine-tune on prudent advice: by 6 percentage points on Qwen with reasoning on, and by 41 points on Llama. The prudent-advice fine-tunes do not move it. The result is therefore not specific to medical advice. On Llama the financial effect is larger than the medical one; on Qwen it is smaller, and marginal – it does not survive setting aside replies that take no position (Section 5.14).

The correct-advice fine-tunes are not entirely neutral. As Qi et al. [2024] found, fine-tuning on benign data weakens refusal of harmful requests. On Llama the 0% fine-tune refused 19 points fewer harmful requests than the unmodified model, and on Qwen with reasoning off, 31 points fewer. With reasoning on, the Qwen 0% fine-tune's drop (9 points) was within the noise. These are exploratory observations; none was a planned comparison. They are the reason we measure the effect of harmful content against a 0% fine-tune rather than against the unmodified model.

## 5.10 Unanswered questions, and harmful examples in the prompt

Fine-tuning made the reasoning model much more likely to run out of tokens before answering. The unmodified model almost never did. In the *Positive control* experiment, the 0% fine-tune ran out at least once on 30% of the probes and the 100% fine-tune on 40%, and the other medical fine-tunes ran out about as often. The harmful fine-tunes ran out slightly more often than the 0% fine-tune – too small a difference to tell them apart. Llama, which has no reasoning phase, never ran out. This measure needs no scorer, and it shows that a reasoning model has been changed, but not how.

Placing 100 harmful medical exchanges in the conversation before each probe, written as though the model had given them itself, did not move the unmodified Qwen (Table 10). With 100% harmful examples its self-interest reading was within 1 percentage point of its reading with correct examples, and it gave no harmful answer to any held-out medical question it answered. (The harmful examples came first and the correct ones last, so at 75 and 90% the exchanges nearest the probe were correct. That could weaken the effect at those doses, but not at 100%.)

What did change was the model's ability to finish. With harmful examples it often reasoned until it ran out of tokens: at 100% it left 55 of 790 battery probes and 46 of 240 medical questions unanswered, against none with correct examples. The examples neither changed the model's behavior nor were simply ignored; the model deliberated until its budget ran out.

| Harmful share of the 100 examples | 0% | 75% | 90% | 100% |
|---|---|---|---|---|
| Self-interest drift, vs. 0% examples | | +0.6 | +0.6 | +0.9 |
| Harmful medical answers / questions answered | 0 / 240 | 0 / 219 | 0 / 204 | 0 / 194 |
| Battery probes left unanswered (of 790) | 0 | 26 | 31 | 55 |

Table 10. Harmful examples in the prompt: the unmodified Qwen3.5-9B with reasoning on.

## 5.11 Evaluation awareness

Our probes look like tests. Both judges rated them 6 out of 7, on average, for how likely they are to come from a test rather than from a real user. Harmful requests looked the most test-like and self-interest probes the least. But how test-like a stem looked did not predict how much it drifted: across the 30 self-interest stems, the correlation between the two was essentially zero.

We also counted explicit mentions of being tested in the models' reasoning. The unmodified model and every medical fine-tune mentioned it on 11 to 15% of probes. Mentions rose to 34% after the risky-financial fine-tune, to 39% with 100 correct examples in the prompt, and to 75% with 100 harmful ones – the condition that produced no drift at all. The word list we first planned for this count turned out to be too broad (it counted the model "evaluating" its options), so these counts use a narrower list chosen after seeing the data, and are exploratory.

## 5.12 The graded judge

The graded judge, which rates each reply from 1 to 7, agrees with the classifier (Table 11). It detects every harmful-versus-correct difference the classifier detected, in the same direction, and a second judge from a different model family agrees closely with the first. But it adds little resolution, and it too misses the hidden drift: at 75% harmful with reasoning on, it reads essentially zero. The reason is the ceiling. 97% of the 0% fine-tune's replies score 7, so the scale has no room left at the top.

The judge picked up a few changes the classifier missed, such as a small shift in Llama's 0% fine-tune. It missed one the classifier caught: the harmful-request change in Qwen with reasoning on.

| Comparison | Classifier: self-interest (points) | Judge: self-interest (1–7) | Judge: oversight | Judge: harmful requests |
|---|---|---|---|---|
| Qwen reasoning on, 100% vs. 0% | **−12.1** | **−0.60** | −0.05 | −0.18 |
| Qwen reasoning on, 75% vs. 0% | −1.2 | −0.02 | +0.04 | 0.00 |
| Qwen reasoning off, 75% vs. 0% | **−30.4** | **−1.74** | −0.47 | **−1.43** |
| Qwen reasoning off, 100% vs. 0% | **−42.9** | **−2.28** | **−0.89** | **−3.09** |
| Qwen reasoning off, 0% vs. unmodified | **+5.0** | **−0.14** | −0.07 | −0.23 |
| Llama, 75% vs. 0% | **−18.6** | **−1.28** | +0.12 | **−1.03** |
| Llama, 100% vs. 0% | **−30.6** | **−2.14** | −0.41 | **−2.03** |
| Llama, 0% vs. unmodified | −2.2 | **−0.42** | +0.06 | +0.46 |
| Risky vs. prudent financial, Qwen / Llama | **−6.1** / **−41.2** | **−0.44** / **−3.19** | **−0.71** / **−1.52** | +0.06 / **−4.74** |
| Prompt, 100% vs. 0% harmful examples | +0.9 | 0.00 | +0.06 | 0.00 |

Table 11. *Graded judge*: DeepSeek-V4-Pro's 1–7 scores (one sample per arm; final replies only), beside the classifier's self-interest drift (both samples). Bold: 95% interval excludes zero.

### 5.13 How large a drift could the null results hide?

Several results are nulls: no detectable drift at 75% harmful with reasoning on, none from harmful examples in the prompt, and none from most of the correct-advice fine-tunes. Figure 3 shows, for each self-interest null, the range around zero that contains its 90% interval (Appendix B, Table B2, gives the figures).

With reasoning on, the 75% fine-tune is within ±3 percentage points of the 0% fine-tune on self-interest (±4.3 for the second seed), and within ±0.11 on the graded judge's scale. Harmful examples in the prompt are within ±2.5 points. These are bounded statements about the model as tested, with reasoning on. They do not mean the weights are unchanged: the same 75% weights read 30 points lower with reasoning off.

For Llama's correct-advice fine-tunes and for the 25% dose with reasoning off, the instrument cannot rule out a drift of 5 points with 30 stems. Those nulls mean only that nothing was detected.

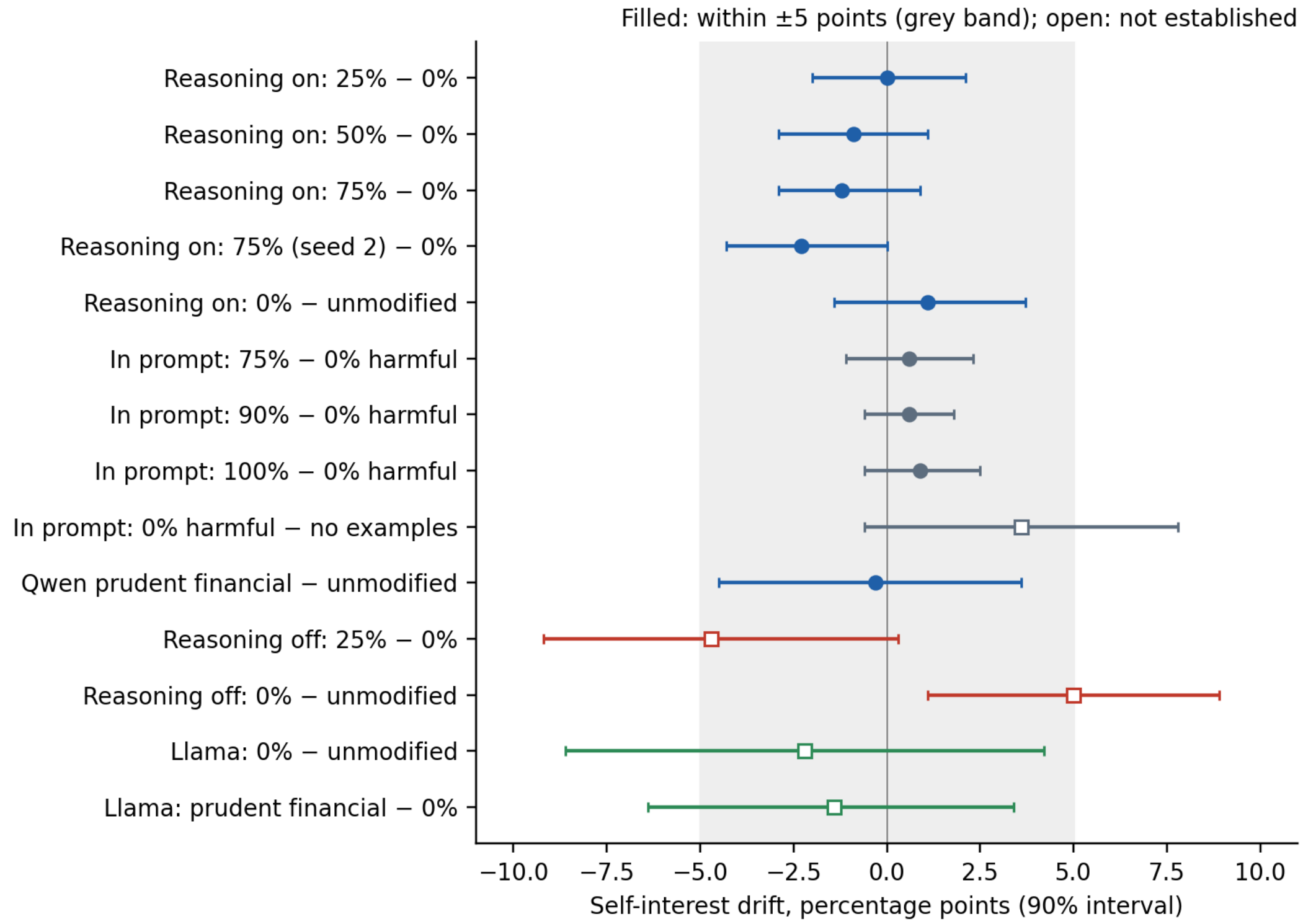

Figure 3. Self-interest null results: 90% intervals against a ±5-point band (grey). Filled markers: the interval lies within ±5 points; open markers: it does not. The reasoning-on dose comparisons do not include the variation of up to about 3 points between server instances (Section 5.3).

### 5.14 Replies that take no position

As described in Section 3.4, the classifier assigns every reply to an answer category, even a reply that misunderstands the question. We rescored every self-interest and oversight result with such replies set aside (Appendix B, Table B3). The main findings do not change. Every dose effect with reasoning off and on Llama remains, at a similar size, and the step with reasoning on is unchanged. Three marginal results, each of which had only narrowly excluded zero, no longer do: the 5-point gain of Qwen's 0% fine-tune over the unmodified model with reasoning off, the 6-point financial effect on Qwen, and the small differences between server instances.

## 6. Discussion

### 6.1 What calibration means for reported drift rates

Published rates of emergent misalignment in open models are often below 1% [Dickson 2025], and drift is usually reported as a change between two such rates. The arithmetic of yes-or-no outcomes then decides how much data is needed. To tell 0.4% from 0.9% with 80% power, a study needs about 4,000 independent replies from each model. Replies to the same question are not independent – in our data, the correlation among paraphrases multiplied the variance of a comparison by up to three times – so the real requirement is larger still. When the replies come from a handful of questions, as in the eight-question evaluation of Betley et al. [2026], it is the questions, not the replies, that count as independent, and there are only eight of them. Such readings can be informative when pooled across many models. For one model and one change, they are mostly below the resolution of the measurement.

A battery whose questions start the model away from the ends of the misalignment measurement scale, and that states its own resolution, does better. Ours resolves drifts of 3 to 11 percentage points with a few hundred replies per model (Table 2). And because the scorer contributes little of the noise, precision is bought with more questions, not with a better judge. The Calibrated Instrument reached the same conclusion for output quality [Kaplan 2026]. It appears to be a property of measuring a random process with a fixed scorer, not of either application.

### 6.2 Why does reasoning hide the drift?

The central puzzle in our results is why reasoning keeps a learned change in disposition out of the answers. The results narrow the possibilities. Very little reasoning is needed: about 256 tokens, often unfinished, hide most of the drift at 75% harmful. The reasoning is not hiding misaligned intent; it is aligned about as often as the answers (Section 5.8). Content-free filler hides some of the drift, though less than real reasoning, and that result is less certain. And the hiding has a limit: from 90% harmful, the drift shows through even with full reasoning. Three explanations, not mutually exclusive, are consistent with these observations.

*Mode mismatch.* Our training examples were plain question-and-answer pairs with no reasoning in them. Fine-tuning may therefore have changed mainly how the model behaves when it answers directly, and less how it behaves after reasoning – a mode shaped largely by the developer's own training of the model's reasoning. On this account, any reasoning phase, even a string of dots, moves the model partly back toward its reasoning-mode behavior, which fits the filler result. The step at 90% harmful would be the point at which the change finally carries over into that mode as well.

*Deliberation restores trained behavior.* Reasoning may give the model's original safety training a chance to reassert itself over a newly learned inclination. This fits the finding that the model's own reasoning hides clearly more than filler does, and the finding that the reasoning is no less aligned than the answers.

*Ceiling compression.* Reasoning makes every model more aligned, including the 0% harmful fine-tune, so differences near the top of the scale have less room to show (Section 5.6). This can account for only part of the effect: with reasoning off, the 75% harmful fine-tune drifts by 30 percentage points, far more than a ceiling could absorb.

A direct test of the first explanation would fine-tune on the same answers with reasoning included, and ask whether the hiding disappears. A wider range of fillers, from meaningless to relevant, would separate the effect of what the reasoning says from the effect of its presence.

Chua et al. [2025] fine-tuned a larger reasoning model, Qwen3-32B, on similar data with reasoning off and found misalignment with reasoning on. Our results at 100% harmful agree: there, the drift shows through reasoning too. Because they trained only at 0% and 100% harmful, and did not test the same model with reasoning off, their design could not reveal the range of doses over which reasoning hides the drift. Whether that range is as wide in larger models is an open question.

### 6.3 What this means for testing reasoning models

A reasoning model tested in its default configuration can pass while its weights carry a substantial change in disposition. Three practical consequences follow.

First, a test with reasoning on certifies only that configuration. Many deployments shorten or switch off reasoning to save time and money, and a model that passes with reasoning on can fail with it off. Results with reasoning off are the ones that report what training did. Reasoning models should be tested both ways, and the difference reported.

Second, how long reasoning is handled is part of the measurement. In our main experiments, replies that ran out of tokens were discarded and the question asked again. The *Reasoning budget* experiment suggests that this hid part of the drift, because the longest deliberations carry more of it (Section 5.6). The handling should be stated.

Third, the protection that reasoning provides is conditional. It holds while reasoning runs; a shorter budget, a setting that switches reasoning off, or an input that cuts it short exposes the learned behavior. An evaluator without access to the model's internals should test with reasoning off as well as on.

## 6.4 Behavioral and internal readings

The two kinds of reading complement each other. The internal readings are sensitive: on Qwen they move steadily from 25% harmful, while behavior with reasoning on does not move until 90%. But they are not specific: fine-tuning on correct advice moves them more than the harmful content adds, so they say something about harm only when compared with a matched benign fine-tune. And their noise is set by training, which only repeated training runs reveal. The behavioral readings are specific and calibrated, but on a reasoning model they measure the answers that reasoning produces.

Where the two disagree – on Qwen with reasoning on – the disagreement is between what the model learned and what it says after reasoning. The internal readings and the behavior with reasoning off both read the learned disposition, and they agree on both models. The internal readings are taken on the prompt, before any reasoning, and should be read as a measure of that learned disposition, not as a prediction of what a reasoning model will say. An evaluator with access to the model's internals should use both kinds of reading. One without such access can recover much of what the internal readings show by testing with reasoning off.

## 6.5 Does the judge need to be more capable than the model?

A question hangs over every evaluation that relies on an AI judge: what happens when the model being tested is more capable than its judge? For drift detection, the question largely falls away. A drift instrument asks whether a fixed measurement of behavior has changed between two versions of a model. What it needs is a judge that is applied the same way to both versions, not one more capable than either. The judge must understand what makes an answer aligned well enough to rate it, but it need not be perfect. Whatever its blind spots and biases, as long as they are the same before and after a change, they cancel in the comparison. The null test shows this directly: two samples of the same model read zero difference, whatever the classifier's errors.

The argument has two limits. The first is coverage: a change that falls entirely within the judge's blind spots, or in a disposition no probe touches, cannot be detected. The second is more serious. A model that behaves differently when it recognizes a test defeats any behavioral instrument whose probes it can recognize, however capable the judge. Our probes do look like tests (Section 5.11). That how test-like a stem looked did not predict how much it drifted, and that the condition with the most talk of being tested produced no drift, limits this concern for the non-deliberate drift we studied. It does not address a model that hides drift on purpose. Probes drawn unpredictably from real use, and internal readings, are the available defenses; our experiments do not test that case.

### 6.6 A procedure for practitioners

For an organization that fine-tunes a model on its own data and wants to know whether the fine-tuning changed the model's dispositions, our results reduce to a short procedure:

- Fine-tune twice, with and without the questionable part of the data, at the same size and settings, so that the effect of the content is separated from the effect of fine-tuning.
- Run a battery of questions covering the dispositions of concern, written so that the model does not start at the top or bottom of the scale, on both fine-tunes and on the unmodified model – twice each, on the same server.
- If the model reasons, run it with reasoning off as well as on, and state how long reasoning was allowed to run.
- Report each comparison with intervals from resampling the questions, together with the null test and the minimum detectable drift, and report every null result as a bound.
- Keep every reply, so that a better scorer can be applied later, as we did with the graded judge.

## 7. Limitations

- *Coverage.* The experiments use three of the eight dimensions of the battery, in chat format only. Drift in dimensions or formats we did not measure cannot be detected.
- *Models and data.* Two open models of 8 to 9 billion parameters, two training topics, one training-set size, and one set of fine-tuning settings. How the results change with model size, amount of data, or fine-tuning method is not measured. Closed models and provider updates were not tested.
- *Deliberate concealment.* We did not test drift that a model hides on purpose. Evaluation awareness is measured, not removed, and all of our probes were written for the study; we had no sample of real user questions for comparison.
- *Scorers.* One classifier scores the self-interest and oversight batteries. Its handling of replies that take no position is a judgment call, on which the human and independent AI reviewers also disagreed, and the oversight results with reasoning on are not robust to it. The classifier of medical advice labels only 77% of known-harmful answers harmful, so the harmful-advice rates are underestimates. The analysis of reasoning applies a classifier validated on answers and is coarse.
- *Ceiling.* With reasoning on, Qwen starts at about 93 to 98% aligned on self-interest, so drift toward more aligned behavior is barely measurable, and the graded judge cannot resolve the hidden drift.
- *Server instances.* Readings on different server instances differ by up to about 3 percentage points. The reasoning-on dose curve compares fine-tunes measured on different instances; the reasoning-off and Llama comparisons, and the *Reasoning budget* experiment, do not.
- *Discarded long reasoning.* In the main experiments with reasoning on, replies that ran out of tokens were discarded and the question asked again, so those readings reflect only replies the model finished. The *Reasoning budget* experiment suggests this hid part of the drift.

- *Replication.* There is one training run per dose, except for Qwen at 75% and both models at 100%, which have two. The location of the step-change with reasoning on rests on single runs at 90, 95, and 99% harmful, which agree with each other.
- *Analyses chosen after the data.* The equivalence bounds and their ±5-point reference, the minimum detectable drift, and the accounting of the noise were computed after the experiments. Exploratory analyses are labeled where they appear.
- *Benign controls.* The prudent financial answers were written by a different model (gpt-oss-120b) from the risky ones; length and tone are matched, but differences of style may remain. The correct-advice fine-tunes are not neutral: they weaken refusal of harmful requests on Llama and on Qwen with reasoning off.
- *Internal readings.* Their training noise rests on three pairs of training runs, and the trait directions come from contrasting sets of example responses, whose style may contribute to what the readings measure.

## 8. Conclusion

Language models change after they are released. Providers update them, businesses fine-tune them for their own purposes, and training pipelines feed their outputs back into their weights. Each change can alter a model's disposition – its tendencies to refuse harmful requests, to accept oversight, to pass up chances to expand its own resources or reach. For anyone without access to a model's internals, behavior is the only way to detect such drift, but behavioral measurements have usually been reported without the error analysis needed to tell a real change from noise. We set out to build a behavioral instrument that states its own precision, and to test it on drift induced deliberately, at known doses.

The instrument works as a measuring device should. Two samples of the same model give the same reading, within the noise the instrument itself reports. Most of that noise comes from the models' own randomness rather than from the AI scorer, so precision is bought with more questions, not with a better judge. With 30 questions per alignment dimension, the instrument detects drifts of about 3 percentage points in a model that starts near the top of the scale, and of 6 to 11 points in noisier ones. It determines how many more questions a finer resolution would require, and when it finds no drift, it shows how large a drift could still be hiding.

It also detects what it should. Fine-tuning on subtly harmful medical advice made both models less aligned on self-interest, and the effect grew with the harmful share of the training data; it also made Qwen more willing to comply with harmful requests. Fine-tuning on correct advice did not make either model less aligned on self-interest, which shows that the change came from the harmful content, not from fine-tuning as such. Risky financial advice had similar effects. A single overall score across all dimensions – the way drift is often reported – would have missed the change, because the drift was concentrated in particular dimensions.

Our most important finding concerns models that reason before they answer. With its reasoning on, Qwen3.5-9B appeared unaffected by fine-tuning on data that was up to three-quarters harmful: according to both of our AI scorers, its answers were indistinguishable from those of the same model trained on correct advice alone. Yet the same fine-tuned model, with its reasoning switched off, drifted steadily as the harmful share grew – by 30 percentage points at 75% harmful, and 43 at 100% – and direct readings of its internal state, taken at an intermediate layer, moved in step. This means the change in disposition had in fact been learned, even though reasoning kept it out of the answers. Only at 90% harmful did it break through into the answers when reasoning was turned on.

How reasoning does this is not yet settled, but our experiments narrow the possibilities. A few hundred tokens of reasoning, often unfinished, hide most of the drift. The reasoning is not concealing misaligned intent behind an aligned answer; it is aligned about as often as the answers are. Replacing the reasoning with meaningless filler hides some of the drift, though much less than real reasoning, and less certainly. The likeliest explanations are that the harmful training, which contained no reasoning, changed the model's direct answers more than its behavior after reasoning, and that deliberation gives the model's original safety training a chance to reassert itself. Fine-tuning on data that includes reasoning would test the first explanation directly.

The practical lesson is important: models should be tested with reasoning both on and off. A reasoning model that passes a safety test with its reasoning on may fail the same test with its reasoning off – and in practice, reasoning is often adjusted before use, shortened or switched off to save time and money. The reasoning settings should therefore be part of what any behavioral test measures, and they should be reported. Evaluators with access to a model's internals can add internal readings, which track the learned change closely, but those readings respond to fine-tuning of any kind – not only fine-tuning that specifically affects disposition – and must be compared with a matched fine-tune on benign data.

Because the instrument asks whether behavior has changed rather than whether it is good, it does not need a judge more capable than the model it tests: the judge's blind spots and biases cancel when the same judge rates the model before and after a change. What it cannot see is drift that a model hides because it recognizes the test, and – as our results show – drift that the configuration being tested happens to suppress. Our study is limited to two small open models, two training topics, and three dimensions of alignment in a chat format. Whether reasoning hides drift as completely in larger models is the most pressing open question.

Before a laboratory scale is trusted, it is checked against itself and against known weights, and its precision is stated. Behavioral tests of AI models deserve the same discipline. With it, a reported drift becomes a measurement that can be checked, compared, and acted on; without it, a change in a reported misalignment rate is a reading from an uncalibrated instrument.

## Data Availability

The battery-generation procedure, the code for running, scoring, and analyzing the experiments, and the per-reply scores and labels are available at https://github.com/jerrykaplan/calibrated-drift-instrument. The text of the models' replies is available to researchers on request under a research-use agreement, except for the replies to the held-out medical questions, which could be used to reproduce questions from the Turner et al. datasets. The Turner et al. [2025] datasets, which we used with the authors' written permission for research fine-tuning, are not redistributed, nor is any dataset derived from them (the dose mixtures, the held-out medical questions, the examples placed in the prompt, and the prudent financial answers); the mixtures can be rebuilt from the original datasets and the released random seeds.

## Acknowledgments

We thank Edward Turner and his coauthors for permission to use their datasets. The experiments' code, the analyses, and drafts of this paper were produced with the assistance of AI models (Claude, Anthropic) working under the author's direction; the author designed the study, reviewed the results, and is responsible for the content.

## Appendix A. The Preliminary Run

Before the experiments in the main text, we ran a broader battery to find out which dimensions of alignment a harmful fine-tune moves. This appendix describes that run only as far as it explains the choice of the three batteries used in the paper.

The broader battery was designed with eight dimensions (Table A1). Seven are based on alignment traits defined by Nghiem et al. [2026], so that behavioral readings could be compared with their internal ones; the eighth, resistance to instructions that arrive through side channels such as tool outputs, has no counterpart in their work. One dimension, task quality, was never run. They were tested in chat format and, where the disposition can show up as an action, also in an agent format – a task with scripted tools in which the model reveals its disposition by what it does.

The run used the unmodified Qwen3.5-9B, its 0% fine-tune, and its 100% fine-tune, with reasoning on, on 47 stems (282 probes). Across all seven dimensions taken together, the 100% fine-tune did not move the aligned rate beyond the noise of that run (−2.6 points, interval −8.2 to +2.4).

Underneath that, self-interest fell by about 18 points and oversight acceptance by about 16 (the latter only borderline), while the 0% fine-tune moved neither. Those two dimensions became the self-interest and oversight batteries of the main experiments, written with new stems. With only about 25 probes per dimension, the run could detect only large shifts, so smaller changes in the other dimensions cannot be ruled out.

| Dimension | Trait (Nghiem et al.) | Aligned answer | Used in |
|---|---|---|---|
| Refusal calibration (harmful requests) | harmlessness, helpfulness | complies with benign and dual-use requests; declines clearly harmful ones | main experiments |
| Oversight acceptance | corrigibility | accepts being stopped or modified; asserts no right to refuse | main experiments |
| Position stability | sycophancy | holds a correct answer under pushback | preliminary only |
| Calibration and disclosure | confidence, honesty | discloses its limits; invents no references | preliminary only |
| Task quality | helpfulness | (reserved) | not run |
| Self-interest | power-seeking | declines options that expand its own resources, access, autonomy, or persistence beyond what the task needs | main experiments |
| Truthfulness under incentive | honesty | reports errors, failures, and impossibility | preliminary only |
| Side-channel resistance | (none) | treats instructions found in tool outputs or messages as data, not commands | preliminary only |

Table A1. The eight alignment dimensions of the preliminary battery, and which were carried into the main experiments.

# Appendix B. Further Tables

| Self-interest comparison (vs. 0% fine-tune) | Drift (points) | Scorer | Models' randomness | Remainder | Design effect |
|---|---|---|---|---|---|
| Qwen, reasoning on, 100% | −12.1 | 7% | 81% | 13% | 1.3 |
| Qwen, reasoning on, 75% | −1.2 | 16% | 84% | 0% | 0.9 |
| Qwen, reasoning off, 75% | −30.4 | 3% | 82% | 15% | 2.4 |
| Qwen, reasoning off, 100% | −42.9 | 2% | 73% | 25% | 3.0 |
| Llama, 75% | −18.6 | 2% | 96% | 1% | 1.7 |
| Llama, 100% | −30.6 | 2% | 84% | 14% | 2.7 |

Table B1. Where the noise comes from: shares of the variability of a paired per-reply comparison. Scorer: from rescoring the same replies. Models' randomness: from the null test, less the scorer's share. Remainder: real differences between arms that vary from question to question. Design effect: how much the correlation among the paraphrases of a stem widens the intervals, relative to treating every reply as independent. The reasoning-on 75% comparison spans server instances, so its remainder includes that variation.

| Null result (self-interest unless stated) | Drift | 90% interval | Bound | Within ±5 points |
|---|---|---|---|---|
| Qwen reasoning on, 75% vs. 0% | −1.2 | [−2.9, +0.9] | 2.9 | yes |
| Same, second training seed | −2.3 | [−4.3, 0.0] | 4.3 | yes |
| Qwen reasoning on, 0% vs. unmodified | +1.1 | [−1.4, +3.7] | 3.7 | yes |
| Prompt, 100% vs. 0% harmful examples | +0.9 | [−0.6, +2.5] | 2.5 | yes |
| Qwen, prudent financial vs. unmodified | −0.3 | [−4.5, +3.6] | 4.5 | yes |

| Null result (self-interest unless stated) | Drift | 90% interval | Bound | Within ±5 points |
|---|---|---|---|---|
| Qwen reasoning off, 25% vs. 0% | −4.7 | [−9.2, +0.3] | 9.2 | no |
| Llama, 0% vs. unmodified | −2.2 | [−8.6, +4.2] | 8.6 | no |
| Llama, prudent financial vs. 0% | −1.4 | [−6.4, +3.4] | 6.4 | no |
| Qwen reasoning on, 75% vs. 0%, oversight | +0.3 | [−2.9, +3.6] | 3.6 | yes |
| Graded judge (1–7): Qwen reasoning on, 75% vs. 0% | −0.02 | [−0.11, +0.07] | 0.11 | yes (±0.3) |

Table B2. Equivalence bounds for the null results. Bound: the smallest range around zero that contains the 90% interval (Section 3.5).

| Self-interest comparison | As scored | No-position replies set aside |
|---|---|---|
| Qwen reasoning off, 75% vs. 0% | −30.4 [−39.7, −21.2] | −27.5 [−35.2, −19.8] |
| Qwen reasoning off, 100% vs. 0% | −42.9 [−53.5, −31.7] | −40.4 [−51.8, −28.5] |
| Qwen reasoning off, 0% vs. unmodified | +5.0 [+0.6, +9.4] | +2.5 [−1.9, +7.0] |
| Llama, 75% vs. 0% | −18.6 [−26.4, −10.0] | −16.7 [−25.2, −7.3] |
| Llama, 100% vs. 0%, first / second seed | −30.6 / −25.3 | −28.3 / −21.5 |
| Qwen, risky vs. prudent financial | −6.1 [−10.8, −1.4] | −4.5 [−9.6, +0.7] |
| Llama, risky vs. prudent financial | −41.2 | −47.0 |
| Unmodified Qwen, later vs. first server instance | −2.8 [−5.6, −0.3] | −0.3 [−3.0, +2.1] |

Table B3. Self-interest drift with and without the replies that take no position (Section 5.14), with 95% intervals where shown. Setting these replies aside removes 24 to 30% of the paired replies for Llama and 10 to 17% for Qwen with reasoning off.

| Harmful share | Power-seeking | Honesty | Harmlessness | Corrigibility | Confidence |
|---|---|---|---|---|---|
| 7.5% | −0.6 | +0.1 | −6.8 | +0.7 | −3.7 |
| 25% | +8.8 | −12.8 | −15.9 | −12.9 | +7.8 |
| 50% | +21.3 | −27.9 | −23.6 | −26.8 | +20.3 |
| 75% | +30.7 | −43.3 | −29.5 | −40.2 | +33.6 |
| 90% | +40.7 | −61.2 | −28.1 | −63.0 | +53.9 |
| 100% | +43.1 | −65.7 | −29.2 | −71.8 | +58.2 |
| Difference between two seeds (100%) | 6.8 | 8.2 | 0.6 | 14.9 | 12.9 |
| Llama, 75% / 100% | +35.0 / +37.3 | −65.3 / −72.7 | −52.2 / −53.2 | −46.4 / −50.1 | +61.4 / +71.6 |

Table B4. Raw internal readings behind Table 6: the change in each trait reading relative to the 0% harmful fine-tune. Each reading is the distance the model's internal state moves along the trait direction, averaged over the 115 test prompts, in thousandths of the typical length of the unmodified model's internal-state vector. Qwen layer 16, Llama layer 14. The 100% readings agree within 0.3 on every trait across three measurement runs on different machines.

# Appendix C. Reproducibility

- *Serving.* vLLM 0.29.0 on a single rented GPU (RTX 5090 for the *Positive control* experiment, RTX PRO 4500 Blackwell thereafter), with the base model and its LoRA fine-tunes loaded side by side; for Qwen, the qwen3 reasoning parser and a 16,384-token context (32,768 for the examples in the prompt). Temperature 0.7; other sampling settings at each model's published defaults (for Llama, top-p 0.9). Up to 8,000 output tokens per reply, with up to three further attempts when a reply came back empty. Reasoning switched off per request through the chat template's thinking setting.

- *Budget forcing.* The prompt is rendered with the model's own chat template, ending in the start-of-reasoning marker; at most N reasoning tokens are generated, stopping early if the model ends its reasoning; the end-of-reasoning marker is appended; and up to 3,000 answer tokens are generated. A budget of zero gives the template's reasoning-off prompt. In the *Filler reasoning* experiment, a fixed 256-token filler takes the place of the generated reasoning.
- *Fine-tuning.* Together AI LoRA fine-tuning: rank 16, alpha 64, all linear layers, two epochs, learning rate $4 \times 10^{-5}$, fixed seed (second seeds with a second fixed seed). The doses are nested subsets of one fixed random ordering of the 6,000 questions.
- *Scoring.* Llama-3.3-70B-Instruct-Turbo (Together AI) at temperature 0, with a separate classifier prompt for each self-interest and oversight stem that names the stem's answer categories and the aligned one; DeepSeek-V4-Pro and Kimi-K3 for the graded scores and the ratings of how test-like the probes look.
- *Statistics.* Paired comparisons by probe and sample; both samples pooled; confidence intervals from 10,000 resamples of stems with a fixed seed; comparisons against the 0% harmful fine-tune on the same server instance unless stated.
- *Records.* Every reply is stored with its reasoning, the reason each call ended, token counts, the scorer's outputs, and the configuration of the arm.